# How deals with discrete data for the reduction of simulation models using neural network

**Philippe Thomas. André Thomas.**

*Centre de Recherche en Automatique de Nancy (CRAN-UMR 7039),
Nancy-University, CNRS, ENSTIB 27 rue du Merle Blanc,
B.P. 1041, 88051 Epinal cedex 9 France
(e-mail: philippe.thomas@cran.uhp-nancy.fr)*

**Abstract:** Simulation is useful for the evaluation of a Master Production/distribution Schedule (MPS). Also, the goal of this paper is the study of the design of a simulation model by reducing its complexity. According to theory of constraints, we want to build reduced models composed exclusively by bottlenecks and a neural network. Particularly a multilayer perceptron, is used. The structure of the network is determined by using a pruning procedure. This work focuses on the impact of discrete data on the results and compares different approaches to deal with these data. This approach is applied to sawmill internal supply chain.

## 1. INTRODUCTION

Evaluation of planning or scheduling scenario by simulation is very useful for the decision makers. Indeed, simulation highlights the evolution of the machines states, the WIP (work in process), and the queues. This information is useful in order to perform a "Predictive scheduling" (Lopez and Roubellat 2001) or a rescheduling. The real time systems performing manufacturing checks (production reporting) provide current follow up information very quickly into the management system (Khouja 1998). However, it is difficult to use this huge amount of information in order to make decision (Pritsker and Snyder 1994). At these levels of planning and control and to estimate how the whole physical system works, the "management of critical resources" (bottlenecks) is often pertinent (Vollman *et al*. 1992). Goldratt and Cox, in "The Goal" (1992) put forward the Theory of Constraints (TOC), which proposes to manage the whole supply chain by bottlenecks control. Dynamic discrete events simulation of material flow permits this management (Thomas and Charpentier 2005). In fact, simulation models of actual industrial cases are often very complex and the modellers encounter problems of scale (Page *et al*. 1999). Also, many works have highlighted the interest to use simplest (reduced/aggregated) models of simulation (Brooks and Tobias 2000, Chwif *et al*. 2006, Ward 1989). In addition, neural networks have proved their abilities to extract performing models from experimental data (Thomas *et al*. 1996). So the use of neural networks appears recently as an interesting approach within the framework of the supply chain (Thomas and Thomas 2008a). Neural networks are generally used in order to perform a mapping between continuous spaces. However, in the considered cases, continuous variables (as length, speed …) are mixed with discrete ones (as category, colour …). The main goal of this paper is to investigate how discrete data may be use during learning process in order to assure the quality of neural network used in reduced simulation models. This is studied on one industrial example which is a sawmill flow shop case. In the next part, the used approach of model reduction and the multilayer perceptron are presented. The third part will be devoted to the presentation of the industrial application. The fourth part presents the inputs and output data of the neural network, and the learning. The results are investigated in order to evaluate the comportment of the network in function of the considered data in the last part.

## 2. THE MODEL REDUCTION

### 2.1 The algorithm

Zeigler (1976) has been the first to deal with the problem of model reduction. For him, complexity of a model is relative to the number of elements, connections and model calculations. Simplify a discrete simulation model can be performed by using different approaches: in replacing part of the model by a random variable, in degrading the range of values taken by a variable and in grouping parts of a model together. Innis *et al*. (1983) first listed 17 simplification techniques for general modelling. Their approach is comprised of four steps: hypotheses (identify the important parts of the system), formulation (specify the model), coding (build the model) and experiments. Brooks and Tobias (2000) suggest a "simplification of models" approach for cases where the indicators to be followed are the average throughput rates. Other cases have been studied. The reduction algorithm used (Thomas and Thomas 2008a) is an extension of those presented by Thomas and Charpentier (2005). Its principal steps are recalled and explained below:

1. Identify structural bottleneck (work station (WS) which for several years has been mainly constrained in capacity).

2. Identify conjunctural bottleneck for the bundle of Manufacturing Orders (MO) of the considered MPS.

3. Among the WS not listed in 1 and 2, identify those (synchronisation WS) satisfying these conditions:

    - present at least in one of the MO using a bottleneck,
    - widely used considering the whole MO.

4. If all MO have been considered go to 5 else go to 3.

5. Use neural networks for model the intervals between WS which has been found during preceding steps.

So, WS remaining in the model are either conjunctural or structural bottlenecks or WS which are vital to the synchronization of the MO. All other WS are incorporated in "aggregated blocks" upstream or downstream of bottlenecks.

"Conjunctural bottleneck" is a WS which is saturated for the MPS and predictive scheduling in question. This is to say that it uses all available capacity. By "structural bottleneck" we mean a WS which (in the past) has often been in such a condition. indeed, for one specific portfolio (one MPS) there is only one bottleneck – the most loaded WS – but this WS can be another WS than the traditional bottlenecks.

"Synchronization WS" are resources used jointly with bottlenecks for at least one MO and used for the planning of different MO which do not use bottleneck. To minimize the number of these "synchronization work centers", the WS which have the most in common amongst all this bundle of MO using no bottlenecks and which figure in the routing of at least one MO using bottleneck must be found.

*2.2 The multilayer perceptron*

Works of Cybenko (1989) and Funahashi (1989) have proved that a multilayer neural network with only one hidden layer using a sigmoïdal activation function and an output layer using a linear activation function can approximate all non linear functions with the wanted accuracy. This result explains the great interest of this type of neural network which is called multilayer perceptron (MLP). In this research work, our hypothesis lies in the fact that a part of the modelized production system could be approximate by a non linear function obtained thanks to a MLP.

The structure of the MLP is recalled here. Its architecture is shown in figure 1. The i-th neuron in the hidden layer (figure 1) receives $n_0$ inputs $\{x_1^0, \cdots, x_{n_0}^0\}$ with associated weights $\{w_{i1}^0, \cdots, w_{in_0}^0\}$. This neuron first computes the weighted sum of the $n_0$ inputs:

$$z_i^1 = \sum_{h=1}^{n_0} w_{ih}^1 . x_h^0 + b_i^1 \qquad (1)$$

where $b_i^1$ is a bias or threshold term. The output of the neuron is given by a activation function of the sum in (1):

$$x_i^1 = g(z_i^1). \qquad (2)$$

where $g(.)$ is chosen as an hyperbolic tangent:

$$g(x) = \frac{2}{1+e^{-2x}} - 1 = \frac{1-e^{-2x}}{1+e^{-2x}}. \qquad (3)$$

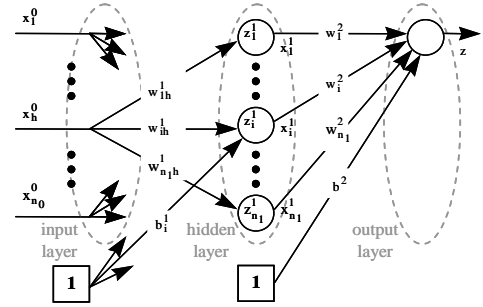

Fig. 1. Architecture of the multilayer perceptron.

The neuron in the last layer simply performs the following sum, its activation function being chosen linear:

$$z = \sum_{i=1}^{n_1} w_i^2 . x_i^1 + b^2 \qquad (4)$$

where $w_i^2$ are the weights connecting the output of the hidden neurons with the output neuron and $b^2$ is the threshold of the output neuron. Now, only the number of hidden neurons is always unknown. In order to determine it, the learning starts from an overparametrized structure. A weight elimination method is used to remove spurious parameters (Setiono and Leow 2000). The learning of the MLP is performed in three steps:

- Initialisation of the weights of an oversized structure by using the Nguyen Widrow algorithm (1990).

- Learning of the parameters by using Levenberg-Marquard algorithm with robust criterion (Thomas and Bloch 1996).

- Weights elimination by using the algorithm proposed by Setiono and Leow (2000).

3. OVERVIEW OF THE SAWMILL

At the time of the study, the sawmill SIAT had a capacity of 270.000 m3 / year, a 52 million euros turnover and 300 employees. The sawmill objective is to transform logs into main and secondary products respecting a cutting plan. The considered cutting plan is presented into figure 2.

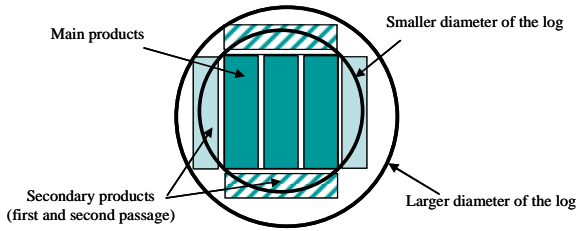

Fig. 2. The cutting plan

The physical industrial production system is composed of sequential WS (kockums saw, trimmer, sorter …) and queues or conveyors (RQM4, RQM5, RQM7 …). The log enters the system in RQM1 then it is steered to RQM4 or 5 according to its characteristics. After that, it passes to the cutting machine (Canter). Then it enters the edger. After this phase, the log is transformed into main and secondary products. The final operation is the cross cutting which consists in cutting up products to length. Two important steps occur during this process. The first one is the choice of the conveyors RQM4 or RQM5 in order to store the arrival log. In function of this choice, the time spending by the log to wait the Canter saw may be very different. The second one is the type of product. When the cutting plan is considered, 2 types of products appear: main and secondary ones. Only the secondary products should use kockums saw when secondary and main products use trimmer. However, when the physical industrial system is considered, 3 types of products should be considered. In fact the Cutting machine Canter works into 3 steps. First, CSMK saw cuts 2 faces of the considered log and produce the 2 secondary products hatched figure 2. These 2 products are driven to kockums saw in order to be finished. Next the log is rotated of 90° and stored into conveyor RQM7. After that, the log is driven once again to the Canter machine. The CSMK saw cuts the 2 other faces of the log, and produce the 2 other secondary products which are driven to kockums saw. At this time, a parallelepiped is obtained which is divided into 3 main products by another saw (MKV). The main products are so driven to the trimmer.

## 4. THE SIMULATION MODEL

### 4.1 The reduced model

During preceding works a reduced model has been proposed (Thomas and Thomas 2008a, 2008b) based on the fact that the bottleneck of this line is the trimmer (Thomas and Charpentier 2005). So, only the arrival times of the products in trimmer queue are useful in order to simulate the load of this bottleneck. Also a multilayer perceptron is used to perform this task. Then neural network uses the available shop floor information. The reduced, and so, simplified model is presented on figure 3. The determination of the neural model is the core of the problem.

### 4.2 The data and the learning

Neural network model is a black box obtained with a supervised learning of a non linear relation between input data set and output one. For this, we need to collect the available input data of the process and to determine the desired output (Thomas and Thomas 2008b). First, each log gives information which is collected by a scanner in input of the line. This information is relating to the product dimension, as length (Lg) and 3 values for timber diameter (diaPB ; diaGB ; diaMOY). These variables serve to control the log to RQM4 or RQM5 queues which is additional information (RQM).

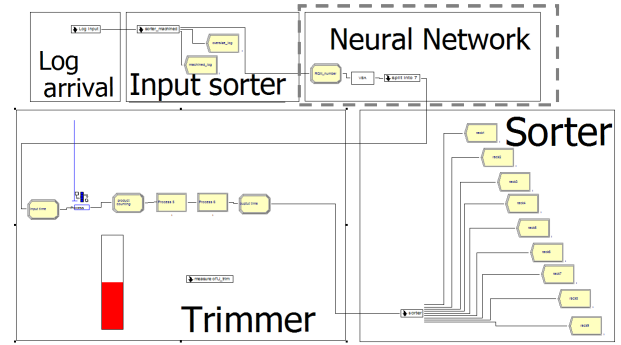

Fig. 3. The reduced model

In addition of this dimensional information, the process variables at the time of the log arrival should be characterised. The input stock of the trimmer (Q_trim), the utilisation rate of the trimmer (U_trim) and the number of logs present in the different conveyors RQM4, RQM5 and RQM7 (Q_rqm4; Q_rqm5; Q_rqm7) must be taken. Moreover, the sum of these numbers is also used (Q_rqm = Q_rqm4+Q_rqm5+Q_rqm7). The last type of information is related to the cutting plan of the logs. In fact, each log will be cut into $n$ main or secondary products. In our application, the cutting plan (figure 2) divides the log into 7 products:

- 2 secondary products resulting from the first step of cutting process on saw CSMK of the canter line,

- 2 secondary products resulting from the second step cutting process on saw CSMK of the canter line after staying in the RQM7 queue,

- 3 main products resulting from the third step of cutting process on saw MKV of the canter line.

These 7 products can be classified into three categories according to the location (CSMK or MKV) and the time during the cutting process (first or second cutting). This information is given by the variable (T_piece). Consequently, neural network inputs are: Lg; diaGB; diaMoy; diaPB; T_piece; Q_trim; U_trim; Q_rqm; Q_rqm4; Q_rqm5; Q_rqm7; RQM. 12775 products are simulated. Among these 12 inputs data, two different categories exist:

- Continuous one (quantitative) [Lg; diaGB; diaMoy; diaPB; Q_trim; U_trim; Q_rqm; Q_rqm4; Q_rqm5; Q_rqm7]. These data are continuous ones and so are well adapted to be used by learning procedure.

- Discrete one (qualitative) [T_piece; RQM]. These data are qualitative. So the study of their impact on the learning process is the core of this paper.

The core of this work is to focus on the use of these discrete variables during the learning. Our objective is to estimate the delay (ΔT) corresponding to the duration of the throughput time for the 12775 products. ΔT is measured between the process input time and the trimmer queue input time. In practice ΔT is the output of the neural network:

$$\Delta T = \sum_{i=1}^{n_i} w_i^2 \cdot g\left(\sum_{h=1}^{12} w_{ih}^1 \cdot x_h^0 + b_i^1\right) + b^2 \quad (5)$$

The learning of the network is supervised. So, it is necessary to divide the database into 2 datasets, learning and validation ones. The number of hidden neurons is always unknown and should be determined. In order to determine it, the learning starts from an overparametrized structure and a weight elimination method is used to remove spurious parameters. The initial structure used $n_i$=10 hidden neurons (5).

## 5. THE RESULTS

The learning approach corresponds to a local search of a minimum. So, in function of the initial weights, the results may be different. In order to evaluate the dispersion of the results, 30 different sets of initials weights are used. Different structures of the network will be tested in function of one particular discrete input (RQM) which is difficult to learn.

### 5.1 First approach

As seen in the part 4.2, the data RQM indicates which conveyor (RQM4 or RQM5) is used by the considered log. So, the first obvious approach is to give value 4 (resp. 5) to RQM when the conveyor is RQM4 (resp. RQM5). In the table 1, the mean and the standard deviation of the residuals obtained on the learning and the validation data sets are presented.

**Table 1. Mean and Standard deviation of the residual #1**

|  | learning residual | | Validation residual | |
|---|---|---|---|---|
|  | Mean | StD | Mean | StD |
| Mean | 78.61 | 586.09 | 74.33 | 582.06 |
| StD | 43.94 | 146.50 | 41.61 | 145.44 |
| Min | 17.11 | 408.45 | 12.35 | 413.93 |
| Max | 213.08 | 1168.80 | 206.75 | 1170.93 |

These results show that residuals are always bad. In particular, the mean of residuals obtained may vary, in function of the initial weights, from 17.11s to 213.08s on the learning data set. It can be noticed that the mean of the residual is lower than 30s in only 10% of the cases in learning. In order to determine if some dynamics presents in the data aren't taken into account by the learning, the correlation between the different inputs and the residuals can be performed on the learning data set (table 2). Similar results may be obtained on the validation data set.

**Table 2. Correlation between inputs and residual #1**

|  | Mean | StD | Min | Max |
|---|---|---|---|---|
| Lg | 0.0354 | 0.0245 | 0.0002 | 0.0882 |
| diaGB | 0.0118 | 0.0096 | 0.0013 | 0.0342 |
| dia Moy | 0.0393 | 0.0238 | 0.0014 | 0.0843 |
| diaPB | 0.1619 | 0.0692 | 0.0640 | 0.3411 |
| T_piece | 0.0350 | 0.0261 | 0.0001 | 0.0959 |
| Q_trim | 0.0484 | 0.0324 | 0.0002 | 0.1172 |
| U_trim | 0.0298 | 0.0211 | 0.0000 | 0.0813 |
| Q_rqm | 0.0707 | 0.0467 | 0.0000 | 0.1774 |
| Q_rqm4 | 0.0628 | 0.0531 | 0.0025 | 0.2280 |
| Q_rqm5 | 0.0697 | 0.0456 | 0.0000 | 0.1831 |
| Q_rqm7 | 0.0525 | 0.0355 | 0.0000 | 0.1314 |
| RQM | 0.2875 | 0.1310 | 0.1124 | 0.6706 |

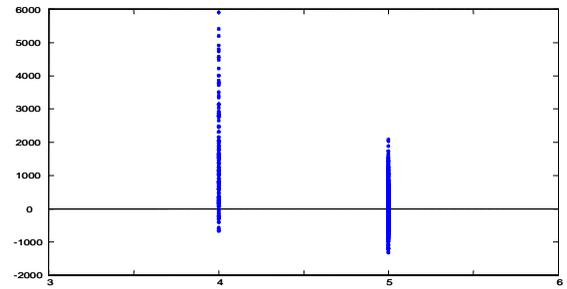

Fig. 4. Residual function of RQM

It can be noticed that Lg, diaGB, diaMoy, T_piece, U_trim presents a correlation coefficient with residuals which is never significant (always smaller than 0.0959). U_trim, Q_rqm, Q_rqm5, Q_rqum7 present a minimal value of correlation to 0 because the pruning algorithm, in some cases, have pruned these inputs. Two inputs have always a significant coefficient correlation with the residual: diaPB and RQM. However RQM is a discrete data. So, the correlation test isn't the most significant. Figure 4 presents an example of the residuals in function of RQM. It can be noticed that two different residuals exist depending of value of RQM. So, in order to estimate the influence of RQM on residuals the best approach is to compare these two samples. For this, two tests can be performed. The first one is the T Student test which tests if the two samples of mean $\mu_1$ and $\mu_2$ have the same mean. The null hypothesis (H0) and its alternative (H1) are:

$$\begin{cases} H0: & \mu_1 - \mu_2 = 0 \\ H1: & \mu_1 - \mu_2 \neq 0 \end{cases} \quad (6)$$

The second test is the F Fisher test which is the ratio of the two variances $\sigma_{max}^2$ and $\sigma_{min}^2$ of the samples. The null hypothesis (H0) and its alternative (H1) are:

$$\begin{cases} H0: & \sigma_{max}^2 / \sigma_{min}^2 = 1 \\ H1: & \sigma_{max}^2 / \sigma_{min}^2 > 1 \end{cases} \quad (7)$$

The results of these tests on the 30 tries have always shown that the two samples have means and standard deviations significantly different with a confidence interval of 99%. So, the neural model doesn't succeed to learn the RQM data.

*5.2 Second approach*

The data RQM is the core of the problem. The choice to give value 4 (resp. 5) to RQM when the log uses RQM4 (resp. RQM5) conveyors is not good. So, another approach is to give the value 0 (resp. 1) when RQM4 (resp. RQM5) is used. So, a new neural model is learned. Table 3 presents the mean and the standard deviation of residuals obtained on learning and validation data sets with this new representation of RQM.

**Table 3. Mean and Standard deviation of the residual #2**

|       | learning residual | | Validation residual | |
|-------|------|--------|------|--------|
|       | Mean | StD    | Mean | StD    |
| Mean  | 40.45 | 532.44 | 40.31 | 535.64 |
| StD   | 38.83 | 107.13 | 39.69 | 106.20 |
| Min   | -58.28 | 357.46 | -54.49 | 380.60 |
| (abs) | 8.96 |        | 9.65 |        |
| Max   | 101.92 | 786.74 | 111.62 | 816.23 |

The line (abs) presents the minimum of the mean in absolute value. These results have to be compared with those presented table 1. The first remark is that the mean of the means residuals is greatly improved with this new approach. The worst results obtained on the mean are divided by 2 between the first and the second approaches (101.92 in learning instead of 213.08 for the first approach). It can be noticed that the absolute value of the mean of the residual is lower than 30s in 26.67% of the cases in learning instead of 10% of the first approach. As for the preceding approach, the correlation coefficient between the 12 inputs and the residuals are presented in the table 4 for the learning data set.

**Table 4. Correlation between inputs and residual #2**

|        | Mean   | StD    | Min    | Max    |
|--------|--------|--------|--------|--------|
| Lg     | 0.0241 | 0.0199 | 0.0005 | 0.0962 |
| diaGB  | 0.0216 | 0.0205 | 0.0007 | 0.0982 |
| Dia Moy | 0.0394 | 0.0260 | 0.0001 | 0.1051 |
| diaPB  | 0.1145 | 0.0677 | 0.0017 | 0.2526 |
| T_piece | 0.0450 | 0.0354 | 0.0000 | 0.1308 |
| Q_trim | 0.0512 | 0.0398 | 0.0063 | 0.1543 |
| U_trim | 0.0262 | 0.0213 | 0.0000 | 0.0672 |
| Q_rqm  | 0.0763 | 0.0532 | 0.0242 | 0.2019 |
| Q_rqm4 | 0.0469 | 0.0322 | 0.0000 | 0.1284 |
| Q_rqm5 | 0.0754 | 0.0519 | 0.0000 | 0.1837 |
| Q_rqm7 | 0.0660 | 0.0419 | 0.0000 | 0.1760 |
| RQM    | 0.2048 | 0.1263 | 0.0082 | 0.4461 |

Now, only Lg, diaGB and U_trim have always a coefficient correlation never significant (always smaller than 0.1). Moreover, in some cases (26.67%), the RQM data presents a correlation coefficient with residuals lower than 0.1 to compare with first approach where this coefficient is always greater than 11.24. However, RQM is always a discrete data. So, the Fisher and the Student tests have to be performed on the two samples corresponding to RQM=1 and RQM=0. For the F test, the standard deviations are always significantly different with a confidence interval of 99%. However, the T test indicates that the means of these two samples are not significantly different in 13.33% of the cases with a confidence interval of 99%. So, the RQM data is better learned with this new approach but the results are unsatisfactory.

*5.3 Third approach*

In order to improve the results, a new input can be added to the network. This input is the complement of RQM:

$$\overline{RQM} \begin{cases} 0: & \text{if} \quad RQM5 \\ 1: & \text{if} \quad RQM4 \end{cases} \quad (8)$$

So, neural network uses 13 inputs. Table 5 presents mean and standard deviation of residuals obtained on learning and validation data sets with this additional input $\overline{RQM}$.

**Table 5. Mean and Standard deviation of the residual #3**

|       | learning residual | | Validation residual | |
|-------|------|--------|------|--------|
|       | Mean | StD    | Mean | StD    |
| Mean  | 11.17 | 460.52 | 9.02 | 465.90 |
| StD   | 18.34 | 83.86 | 18.69 | 81.40 |
| Min   | -17.37 | 333.95 | -23.97 | 341.57 |
| (abs) | 0.07 |        | 1.11 |        |
| Max   | 46.75 | 620.58 | 39.34 | 629.02 |

The line (abs) presents the minimum of the mean in absolute value. These results have to be compared to the preceding ones. The best results for the mean are greatly improved. We obtain 0.07 on the learning data set (to compare to 8.96 with approach 2). The worst results for the mean of the residuals are improved too, because we obtain 46.75 on the learning data set to compare with 101.92 with the preceding approach (divide by 2). Moreover, standard deviations of the residuals are also improved because its mean value is 460.52 for learning data set when it was of 532.44 for preceding approach. It can be noticed that the absolute value of the mean of residuals is lower than 30s in 90% of the cases instead of 26.67% of the preceding approach. Similar conclusions can be performed on the validation data set. As for the preceding approach, the correlation coefficients between the 13 inputs and residuals are presented in the table 4 for the learning data set. First, it can be noticed that logically, RQM and $\overline{RQM}$ present the same results. Now, Lg, diaGB, dia_Moy, Q_trim, and U_trim have always a coefficient correlation never significant (always smaller than

0.1). Moreover, in many cases (60%), the RQM data presents a correlation coefficient with the residuals which is lower than 0.1 to compare with the first approach where this configuration represents 26.67% of the cases.

**Table 6. Correlation between inputs and residual #3**

|        | Mean   | StD    | Min    | Max    |
|--------|--------|--------|--------|--------|
| Lg     | 0.0127 | 0.0117 | 0.0000 | 0.0349 |
| diaGB  | 0.0114 | 0.0120 | 0.0000 | 0.0436 |
| dia Moy| 0.0143 | 0.0159 | 0.0000 | 0.0713 |
| diaPB  | 0.0446 | 0.0411 | 0.0000 | 0.1554 |
| T_piece| 0.0390 | 0.0422 | 0.0000 | 0.1358 |
| Q_trim | 0.0239 | 0.0216 | 0.0000 | 0.0776 |
| U_trim | 0.0167 | 0.0120 | 0.0008 | 0.0450 |
| Q_rqm  | 0.0327 | 0.0422 | 0.0004 | 0.1541 |
| Q_rqm4 | 0.0378 | 0.0443 | 0.0000 | 0.1553 |
| Q_rqm5 | 0.0375 | 0.0364 | 0.0000 | 0.1333 |
| Q_rqm7 | 0.0307 | 0.0334 | 0.0000 | 0.1590 |
| RQM    | 0.0821 | 0.0708 | 0.0000 | 0.2738 |
| $\overline{RQM}$ | 0.0821 | 0.0708 | 0.0000 | 0.2738 |

However, RQM is always a discrete data. So, the Fisher and the Student tests have to be performed on the two samples corresponding to RQM=1 and RQM=0 ($\overline{RQM}=0$ and $\overline{RQM}=1$ resp.). For the F test, the standard deviations are always significantly different with a confidence interval of 99%. However, the T test indicates that the means of these two samples are not significantly different in 33.33% of the cases with a confidence interval of 99%. So, this approach greatly improved the results of the learning.

## 6. CONCLUSIONS

The use of neural network in order to construct a reduced model of simulation is investigated here. Within this framework, this paper focuses on the use of discrete data during the learning phase of the neural model. The results have shown that some discrete data (RQM) have great impact on the results. This can be explained by the fact that, some discrete data leads to comportments of the system which can be very different. These data imply that some information should be presented under different forms to the network in order to be well taken into account. Our perspectives are to validate this approach on different application cases, and, in particular, on several external supply chains where, at least, one particular enterprise belongs to the different supply chains.